\documentclass[letterpaper, 10 pt, conference]{ieeeconf}  

\IEEEoverridecommandlockouts                              

\usepackage{graphics} 
\usepackage{epsfig} 
\usepackage{amsmath} 
\usepackage{amssymb}  
\usepackage{balance}
\usepackage{algorithm,algorithmic}
\usepackage{multirow}
\usepackage{url}
\usepackage{color}
\newcommand{\todo}[1]{}
\renewcommand{\todo}[1]{{\color{red}Todo: {#1}}}
\usepackage[font=small,labelfont=bf]{caption}

\title{\LARGE \bf
Contact Localization for Robot Arms in Motion without Torque Sensing
}

\author{Jacky Liang$^{1}$, Oliver Kroemer$^{1}$
\thanks{$^{1}$Carnegie Mellon University.
        {\tt\small \{jackyliang, okroemer\}@cmu.edu}}%
}

\begin{document}

\setlength{\textfloatsep}{0.1cm}
\setlength{\floatsep}{0.1cm}

\maketitle
\thispagestyle{empty}
\pagestyle{empty}
\begin{abstract}
Detecting and localizing contacts is essential for robot manipulators to perform contact-rich tasks in unstructured environments.
While robot skins can localize contacts on the surface of robot arms, these sensors are not yet robust or easily accessible.
As such, prior works have explored using proprioceptive observations, such as joint velocities and torques, to perform contact localization.
Many past approaches assume the robot is static during contact incident, a single contact is made at a time, or having access to accurate dynamics models and joint torque sensing.
In this work, we relax these assumptions and propose using Domain Randomization to train a neural network to localize contacts of robot arms in motion without joint torque observations.
Our method uses a novel cylindrical projection encoding of the robot arm surface, which allows the network to use convolution layers to process input features and transposed convolution layers to predict contacts.
The trained network achieves a contact detection accuracy of $91.5$\% and a mean contact localization error of $3.0$cm.
We further demonstrate an application of the contact localization model in an obstacle mapping task, evaluated in both simulation and the real world.
\end{abstract}
\section{INTRODUCTION}

For robot manipulators to robustly operate in unstructured environments, safely interact with humans, and perform contact-rich tasks, they must be able to sense the contacts they make with the external environment.
Indeed, works in tactile sensing have seen many uses in robot manipulation~\cite{li2020review}, including object localization~\cite{koval2017manifold, bauza2019tactile, liang2020hand}, shape completion~\cite{wang20183d, jung2019active, ottenhaus2019visuo}, and active environment exploration~\cite{matsubara2017active, ottenhaus2018active, saund2019blindfolded}.
For robot skins, recent developments demonstrate promising applications in contact estimation~\cite{chavez2018contact, hirai2018tough}, whole-body manipulation~\cite{mittendorfer2015realizing}, compliant control~\cite{bhattacharjee2013tactile, dean2019whole}, safe human-robot interactions~\cite{dean2017tomm, pang2018development}, object classification~\cite{wade2017force, kaboli2018robust, kaboli2018active}, and manipulation in dense clutter~\cite{bhattacharjee2014robotic}.

However, while robot skins is an area under active research and development~\cite{cheng2019comprehensive}, robust and affordable skins that work across a wide variety of robot form factors remain inaccessible.
By contrast, proprioceptive sensing that gives joint angles and velocities, and sometimes estimated torques, are available in most commercial manipulator arms.
As such, prior works have explored using proprioceptive sensors to localize contacts, with proposed methods ranging from model-based optimization~\cite{manuelli2016localizing, benallegue2018model, zwiener2019armcl, wangcontact} to model-free learning~\cite{zwiener2018contact, popov2019real, popov2020transfer}.

Many past works take the perspective of localizing contacts on a static robot arm - the arm is initially set still, then one or more point contacts are established, pushing in the direction normal to the robot mesh.
A typical additional assumption is that the point contact stays in the same position on the robot arm, even if the arm moves as a result of the contact.
While this approach is useful in the context of ``passive" contact incidence, where the the obstacles ``come" to the robot, we argue it is less realistic in the context of ``active" contact incidence, where the robot ``goes" to the obstacle, whether for exploration or manipulation.

\begin{figure}[!t]
    \centering
    \includegraphics[width=0.9\linewidth]{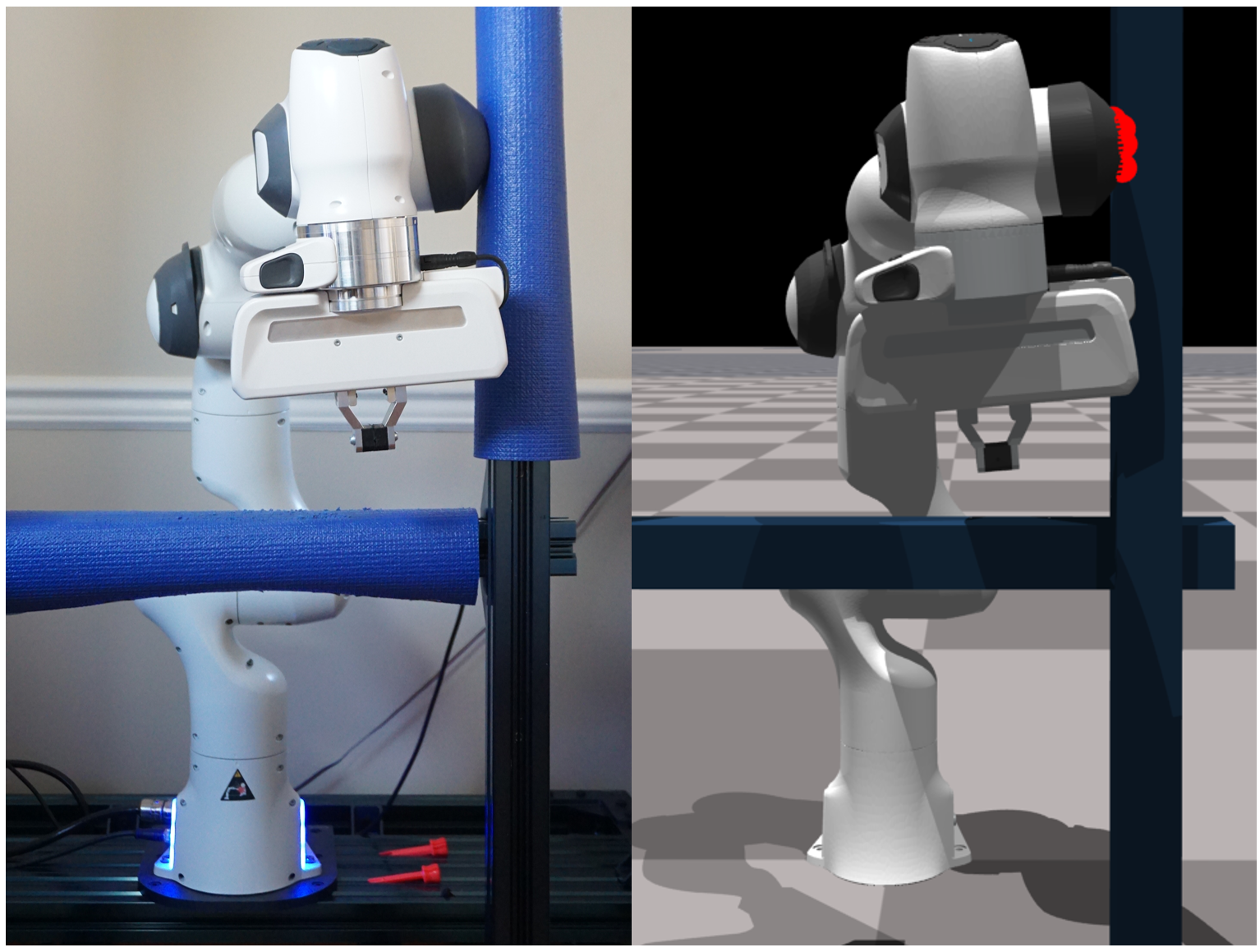}
    \caption{
        A learned neural network contact localization model predicts contact points, visualized in red, on the Franka robot.
        The model is trained in simulation via Domain Randomization, and it does not require torque observations.
        To aid sample complexity and localization performance, the network uses transposed convolution layers to predict contact distance fields in the cylindrical projection space of the mesh surface.
    }
    \label{fig:teaser}
\end{figure}

In this paper, we train a neural network to localize contacts on the surface of a robot arm that is in motion and interacting with obstacles.
The neural network is trained with data collected in simulation, where ground truth contact information can be obtained.
When a robot is in collision with an external obstacle, its proprioceptive responses vary depending on the dynamics model and inertial parameters of the robot.
Because there are mismatches in simulated and real dynamics (sim-to-real gap), we use Domain Randomization~\cite{tobin2017domain} to generate data across a distribution of robot dynamics models.

Furthermore, because of the sim-to-real gap, our network does not rely on torque observations to perform contact localization.
This relaxes the requirement of having accurate force-torque sensing, and it follows from the observation made in~\cite{wangcontact}, that contact points with static obstacles must have zero velocities in the direction of the surface normal.
While the algorithm in~\cite{wangcontact} solves for potential contact locations with joint velocities, our approach uses velocity observations as features to a neural network.

In addition, we use a novel representation to encode both features and contact localization predictions.
This representation projects points on the surface of robot links to cylindrical coordinates, allowing us to represent features and contact points as images.
With this encoding, the model can use convolution layers to efficiently process features of points on link surfaces, such as point-wise link velocities.
We also propose using transposed convolution layers to predict images of contact points, represented as distance fields, on the surface of a robot link.
This is unlike previous works~\cite{zwiener2018contact, popov2019real, popov2020transfer}, which directly classified the contact state of each of these points.
Both transposed convolution and classification-based variants are evaluated, and we demonstrate an application of the learned contact localization networks in an obstacle-mapping experiment, conducted in both simulation and the real world.
See video and supplementary materials at~\url{https://sites.google.com/view/ct-loc}

\section{RELATED WORKS}

Prior works have studied combining a robot's proprioceptive observations and dynamics model to infer the robot's contact state.
The authors of~\cite{manuelli2016localizing} proposed the Contact Particle Filter (CPF), a model-based optimization method that filters for external contacts on a humanoid robot by observing its joint torque residuals - the difference between measured and expected joint torques.
The algorithm approximates solving contact locations as a quadratic program (QP).
For the CPF's observation model, the contact likelihood is proportional to the error of the QP's solution.
For its dynamics model, it assumes that contacts occur at fixed points on and relative to the robot surface - this means the contact point follows the robot link's movements.
While the CPF is efficient and achieves good error rate ($0.4$s per filter step for 3 concurrent contacts, $2$cm localization error), it assumes access to an accurate dynamics model of the robot.

Later works approached contact localization via torque residual observations from a data-driven perspective.
In~\cite{narukawa2017real} the authors devise a real-time collision detection and localization algorithm for a humanoid robot by training a Support Vector Machine (SVM).
A one-class SVM was used for detection, and multi-class SVM for localization.
The work assumes one contact at a time, and the SVMs were trained on real-world data.
While the detection model was trained with data where the robot arm is moving, the localization model assumed static robot configuration - the robot is not moving when the contact occurs.
Furthermore, the localization model predicts only coarse labels - $7$ classes for the entire arm of the humanoid.
Additional contact detection works that focused on locomotion robots improved performance of contact detection, but not localization~\cite{bledt2018contact, rotella2018unsupervised}.

In~\cite{zwiener2018contact} the authors improved the resolution of data-driven contact localization by training machine learning models (Random Forests and Multi-layer Perceptrons) to classify the contact state of $661$ pre-specified points on a 7-DoF Jaco arm's surface.
The work assumes single contacts, that contacts are perpendicular to the robot's surface normals, and that there are no contact torques.
The features of the model consist of a sliding window of joint positions, velocities, accelerations, torques, and linear accelerations, and data was generated in simulation.
The best-performing model achieved a mean localization error of $4$cm with $14$\% False Negative Rate (FNR) (not detecting a contact when there is one), and its inference frequency is higher than $200$Hz, which is much faster than the optimization-based baseline.
Like~\cite{narukawa2017real}, the models also make the static configuration assumption, where the training data consists of $20$ static arm configurations.
The authors' follow-up work~\cite{zwiener2019armcl} proposed a particle filter approach that constrained particles of contact locations to the surface of the robot arm.
This method achieved comparable performance to those in the earlier work, but it had the added capability of handling multiple contacts.

In~\cite{popov2019real} the authors propose a similar a data-driven approach to classify contacts on pre-specified points on a robot arm surface.
Training data were generated in simulation, and like~\cite{zwiener2018contact}, this work also assumes one contact at a time, static robot configuration, and access to joint torque observations.
The follow-up work~\cite{popov2020transfer} improved model performance by training with both simulation and real-world data, achieving $6.4$cm mean localization error.
In addition to detecting contacts, the algorithm in~\cite{cioffi2020data} can also classify between expected and unexpected contacts.
However, the algorithm only gave coarse localization labels - one for the upper arm and one for the lower arm.

Recent works have also explored estimating external contacts without access to joint torque sensors.
The authors of~\cite{benallegue2018model} propose a model-based algorithm that only needs end-effector force sensors and IMUs placed on the robot.
The work also assumes a single contact, but the humanoid robot is moving during contact incident.
In~\cite{wangcontact}, the authors leverage joint velocity measurements to localize contacts.
While the method achieves sub-centimeter localization error, it also assumes single contacts and was applied to a robot with planar kinematics - the links only moved in 2D. 
Our work is related to~\cite{benallegue2018model, wangcontact} in that we do not assume access to joint torque readings.
Like~\cite{zwiener2018contact, popov2019real}, our approach predicts dense contact locations on the surface of 7-DoF robot arms, but our method also allows for predicting multiple contacts and predicting contacts while the robot is in motion.
\section{METHOD}

\begin{figure}[!t]
    \centering
    \vspace{6pt}
    \includegraphics[width=0.9\linewidth]{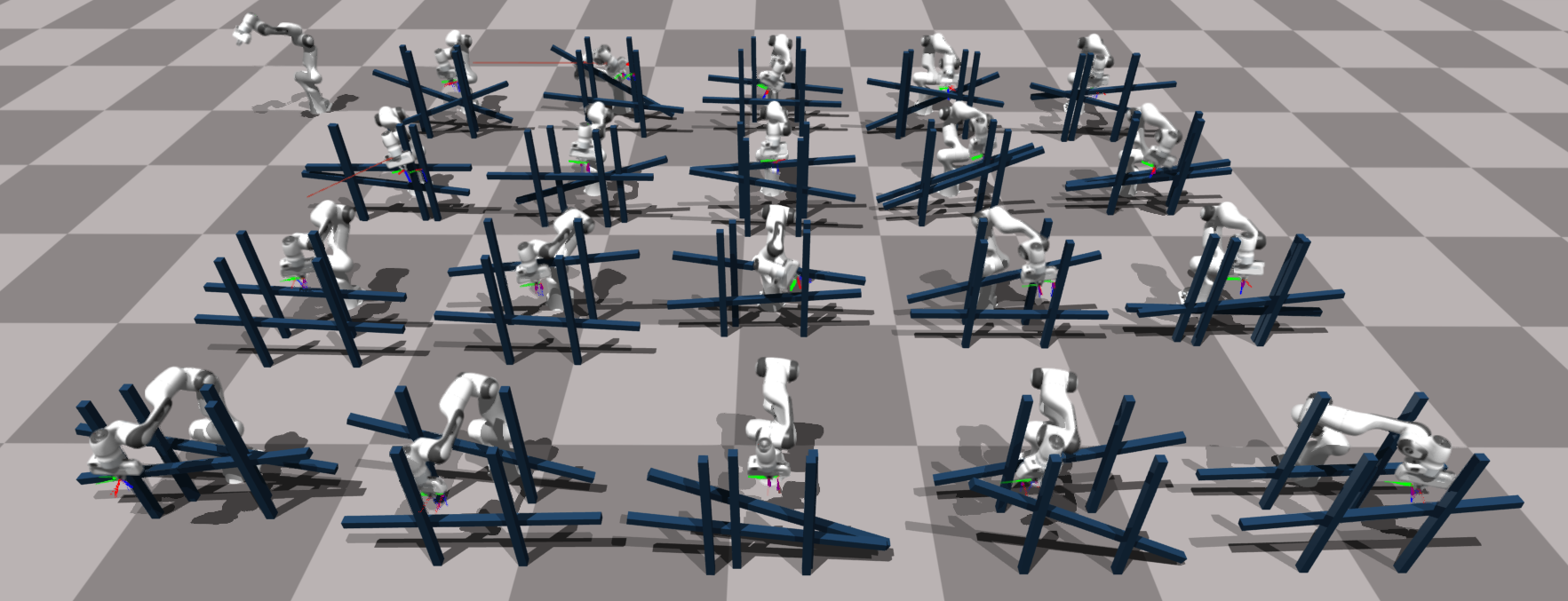}
    \caption{
        Visualization of randomly sampled obstacle environments used for generating contact interaction data. 
    }
    \label{fig:envs}
\end{figure}

\begin{figure*}[!t]
    \centering
    \vspace{3pt}
    \includegraphics[width=0.9\linewidth]{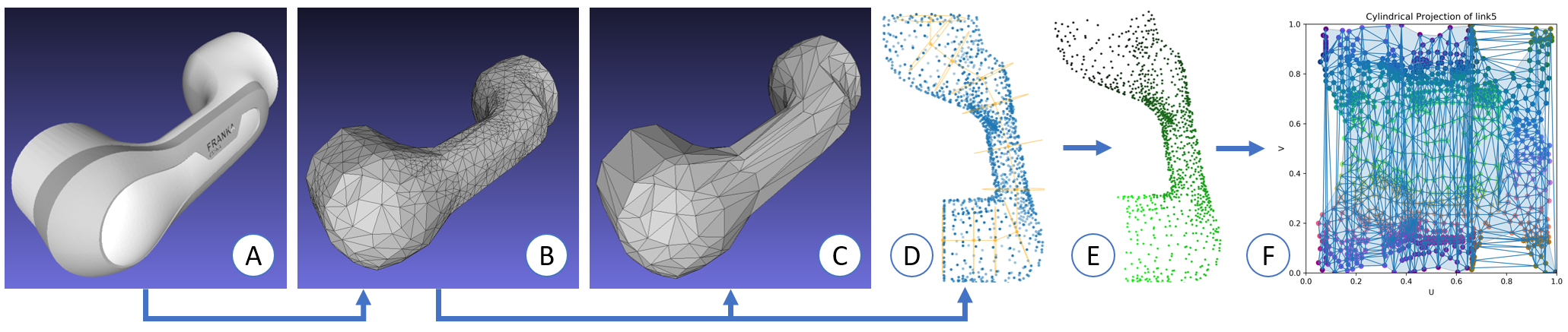}
    \caption{
        Mesh simplification and cylindrical projection pipeline for link 5 on the Panda robot.
        A: Visual mesh provided by Franka Emika.
        B: Projection mesh with lower triangle count, more uniform vertex, and more equilateral triangles obtained via TetWild.
        C: Collision mesh obtained by decimating the projection mesh.
        D: Manually specifying waypoints for a tube (yellow lines and circles) to fit around the projection mesh (blue vertices).
        E: Projecting vertices onto the tube. Here the colors visualize the coordinates along the tube. Greener means the vertex is toward one end of the tube, and black the other.
        F: Normalized cylindrical projection coordinates in the range of $[0, 1]$. 
        The horizontal $U$-axis denotes the coordinate along the tube, and vertical $V$-axis denotes the angular coordinate along the tube's circumference. 
        Each dot is a vertex, and the color of the dot indicates its normal direction. 
        Edges are mesh triangle edges. 
        Blue shaded region denotes the concave hull used to define valid projection interpolation region.
    }
    \vspace{-20pt}
    \label{fig:mesh}
\end{figure*}

We define the contact localization problem as follows:
given a sequence of $T$ observations $o_{1:T}$ sampled at a frequency $f$, detect whether or not each link $l\in[1,\hdots,L]$ of the robot arm is in contact with an obstacle at time $T$, and if it is, also predict the contact locations $c^l_i \in \mathcal{S}_l$.
Here, $c^l_i$ denotes the $i$th predicted contact point of link $l$, which lies on the link's 3D surface manifold $\mathcal{S}_l$, and $L$ is the total number of robot links for which contact localization is performed.
We use $b^l \in \{0, 1\}$ to denote the binary contact state of each link.
We assume the obstacles with which the robot makes contact are stiff, rigid, and stationary.

A neural network is trained to make this prediction.
Training data is collected in simulation with a distribution of robot dynamics models.
There are three important differences between our approach and previous works which also used machine learning to localize contacts for robot arms.
First, our data is generated by having the robot arm interact with obstacles, instead of directly applying contact forces on the arm.
This is important, because it leads to more realistic sequences of robot movements under contact - contact points often slide along the robot arm as it pushes against an obstacle.
Second, we do not use torque observations and do not assume access to precise robot dynamics models, which makes our method more general and applicable to robots without accurate force/torque sensing.
Third, we propose a novel cylindrical projection mapping to represent the robot link surface manifold, which can be used to encode both features and contact localization outputs.

\subsection{Data Generation}

\begin{figure*}[!t]
    \centering
    \vspace{3pt}
    \includegraphics[width=0.9\linewidth]{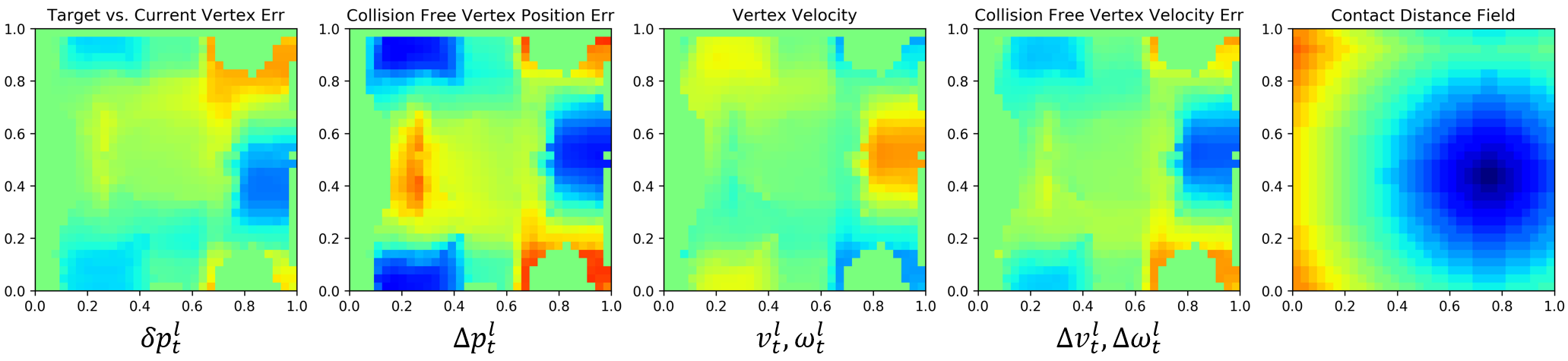}
    \vspace{-5pt}
    \caption{
        Example visualization of features interpolated on the cylindrical projection (left 4) and contact distance field (right most) for link 5 on the Panda arm.
        For the feature images, blue means negative values, green means $0$, and red means positive values.
        For the contact distance field, blue means $0$, and red means the maximum possible distance on the image.
        Note the wrap around in the vertical axis of the contact distance field - this is because the vertical axis corresponds to the angular coordinates of the cylindrical projection.
        For the feature images, pixels outside the valid interpolation region have been zeroed out (see blue shaded region in panel E of Figure~\ref{fig:mesh}).
    }
    \label{fig:proj}
    \vspace{-20pt}
\end{figure*}

We generate the training data by having a robot arm execute exploration trajectories in environments with randomly generated obstacles.
The robot we use is the 7-DoF Franka Emika Panda arms.
We collect features from and make predictions for the last $7$ links on the Franka arm, excluding the first two base links, as they seldom come into contact with external obstacles.
Obstacles in the scene consists of two sets of $3$ beams, $2$ vertical and $1$ across, which are placed in front of the robot.
We randomize the pose of beams, with the across beams attached to the vertical beams.
This results in a diverse set of training environments (Figure~\ref{fig:envs}).

There are two types of exploration trajectories the robot executes.
The first we call random exploration, in which the robot randomly samples a sequence of delta end-effector pose targets and follows through each waypoint with randomly sampled time horizons.
The second we call informed exploration, in which we predefine a sequence of waypoints that are likely to bring the robot in contact with obstacles in the scene, and the robot follows through each one with small amounts of added noise.
Our training dataset contains data from both the random and informed exploration policies, with a $10:1$ ratio respectively.
Details about trajectory generation can be found in the Appendix.

To go to each waypoint, the robot uses min-jerk interpolation and end-effector Cartesian-space impedance control, which converts errors in Cartesian space to torque commands via a spring-damper system.
The simulation uses the Franka dynamics model from~\cite{gaz2019dynamic}, which was fitted on a real Franka robot; this allows realistic impedance control behavior in simulation.
We randomize the inertial parameters of the Franka dynamics model (mass, center of mass, and moment of inertia for each robot link) as well as the gains of the impedance controller for each trajectory.

We collect the following observations: joint angle velocities $\dot{q}_t\in\mathbb{R}^7$, linear velocities of each link $v^l_t\in\mathbb{R}^3$, angular velocities of each link $\omega^l_t\in\mathbb{R}^3$, difference between the current and target joint angles $\delta q_t\in\mathbb{R}^7$, difference between the current and target link poses $\delta p^l_t\in SE(3)$, difference between the current and ``collision-free" joint angles and joint angle velocities $\Delta q\in\mathbb{R}^7, \Delta \dot{q}_t\in\mathbb{R}^7$, and the difference between the current and ``collision-free" link poses and velocities $\Delta p^l_t\in SE(3), \Delta v^l_t\in\mathbb{R}^3, \Delta \omega_t\in\mathbb{R}^3$.
We assume the robot can directly measure its joint angles and joint velocities, from which link velocities can be computed.
The ``collision-free" observations are obtained by running the simulation from the previous step in a separate simulation that has no obstacles.
If the robot is not near an obstacle, then $\Delta p_t$, $\Delta v_t$, and $\Delta \omega_t$ are all zero vectors.
Otherwise, some of these values might be non-zero, and they indicate how much the observed trajectory deviates from the expected trajectory if no obstacles were present.
The target link poses are obtained by running forward kinematics on the desired joints $q_d = q + J^\top p_e$, where $q\in\mathbb{R}^7$ are the current joint angles, $J$ is the analytical Jacobian, and $p_e\in SE(3)$ is the end-effector pose error used by the impedance controller.
We do not use positional observations, like joint angles or end-effector poses.
Avoiding them for training the network ensures that it is not overfitting to the locations of obstacles from the training data.

All training data is generated with Nvidia Isaac Gym~\footnote{\url{https://developer.nvidia.com/isaac-gym}}, a GPU accelerated robotics simulator~\cite{liang2018gpu}.
We set the simulator $dt=0.01$s, and each trajectory runs for $500$ simulator steps, which corresponds to $5$s.
We collect an observation every $4$ simulator steps (so the observations are collected at a frequency of $f=25$Hz), so each trajectory contains $125$ observations.
In total we collect $2800$ simulated trajectories.
The collected trajectories are cleaned by removing all contacts that are less than $1$N in magnitude and last less than $0.1$s.
This resulted in a dataset with about $2$\% of the samples having positive contacts.

\subsection{Link Mesh Processing}

See panels A, B, and C in Figure~\ref{fig:mesh} for the mesh simplification pipeline.
There are three sets of meshes used.
The first are the \textit{visual meshes} obtained from the official Franka repository~\footnote{\url{https://github.com/frankaemika/franka_ros}}.
These are used for rendering and visualization only.
The second set are the \textit{projection meshes}, which are used for cylindrical projections, have lower resolution than the visual meshes.
They also have vertices that are more spatially uniform and triangles that are more equilateral. 
We produce projection meshes by running TetWild~\cite{Hu:2018:TMW:3197517.3201353} on the visual meshes.
Each projection mesh has about $2000$ triangles.
The third set are the \textit{collision meshes}, which are used by the simulation to perform collision checking.
Having even lower resolution than the projection meshes, they are produced by running mesh decimation functions from libigl~\cite{libigl} and MeshLab~\cite{cignoni2008meshlab} on the projection meshes.
Each collision mesh has about $1000$ triangles.

\begin{figure*}[!t]
    \centering
    \vspace{2pt}
    \includegraphics[width=0.8\linewidth]{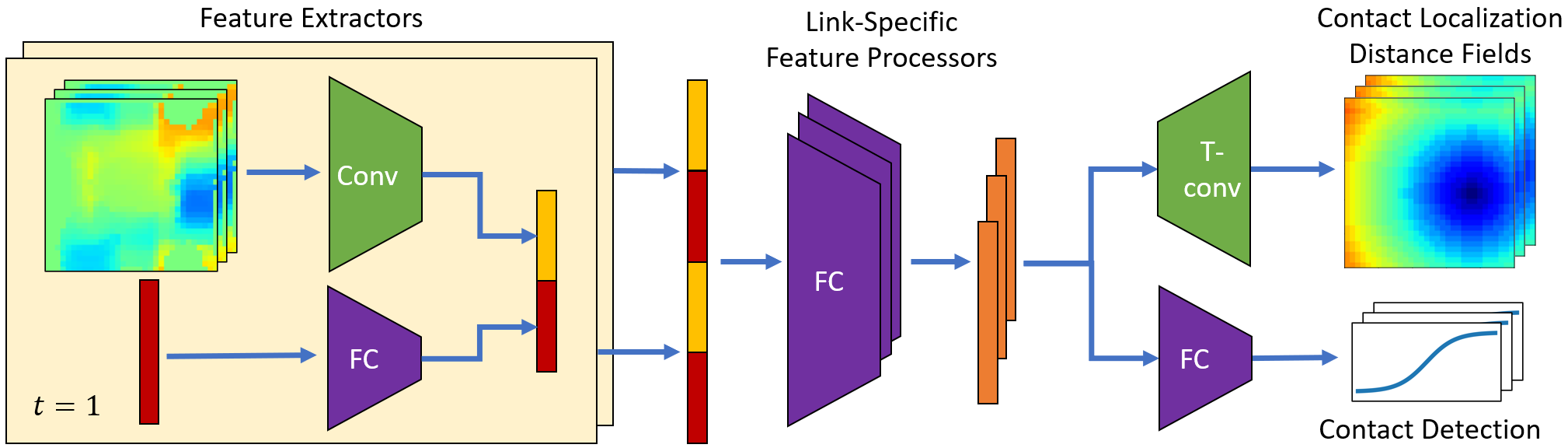}
    \caption{
        Contact Localization Neural Network Architecture.
        The figure visualizes the case for $3$ robot links (denoted by the elements that are repeated thrice) and an observation window of $2$ timesteps (denote by the $2$ repeated light yellow groups on the left).
        In practice we observe features and make predictions for $7$ robot links across a window of $5$ timesteps.
        Conv means convolution layers, FC means fully-connected, and T-conv means transposed convolutions.
        Note that for the inputs to the convolution layers, each square represents feature maps for \textit{one} robot link, which by itself contains $5$ channels.
    }
    \label{fig:network}
    \vspace{-20pt}
\end{figure*}

\subsection{Cylindrical Projections}

See panels D, E, and F in Figure~\ref{fig:mesh} for the cylindrical projection pipeline.
In our approach, obtaining the cylindrical projection representation of a mesh surface means obtaining a mapping from each vertex on the projection mesh to a normalized 2D cylindrical coordinate system.
First, we manually specify a linear-interpolated spline with keypoints that roughly follow the geometry of a link mesh.
This is only done once per robot link and takes about $5$ minutes.
Then, we form a tube along the spline and project vertices on the link to the surface of this tube.
This step can be thought of as contracting a ``sleeve" around a robot link.
After this step, each mesh vertex is assigned two coordinates: $s$ and $\phi$, where $s$ denotes the location of the mesh vertex along the spline, and $\phi$ denote the angle of vertex along the tube circumference.
Lastly, these $(s, \phi)$ coordinates are normalized into $(u, v)$ coordinates, where we map $s \in [s_{\text{min}}, s_{\text{max}}] \rightarrow u \in [0, 1]$ and $\phi\in [0, 2\pi] \rightarrow v \in [0, 1]$.
To map from a mesh vertex to the cylindrical coordinates, we simply read off its precomputed $(u, v)$ values.
To map from a point in the $UV$ space back to a mesh vertex, we use the mesh vertex whose cylindrical coordinates are the closest to the query point in $UV$ space.

We use the cylindrical projection for two purposes - encoding contact distance fields which our trained neural network tries to predict, and encoding link surface features which the neural network uses for prediction.
See Figure~\ref{fig:proj} for examples of both.
To generate the contact distance fields for training, we first find the $UV$ coordinates of the mesh vertices that are in contact, and then generate a distance field image, where the value of each pixel is the distance to the closest contact point in $UV$ space.
We use $32\times 32$ as the image resolution, and it spans the entire $UV$ space of $[0, 1]\times[0, 1]$.
Importantly, distance in the $V$-axis ``wraps around" as it corresponds to angles on a cylinder.
This representation is desirable, because it easily allows encoding multiple contact points, and it provides a smooth and dense supervision signal for training - even points that are not in contact have useful information about the neighboring points that are.
The latter is especially important in reducing sample complexity and improving network performance.

There are $5$ types of features encoded with the cylindrical projections: 1) difference between the current and target link poses $\delta p^l_t$, 2) link velocities $v^l_t$, $\omega^l_t$, 3) difference between the current and collision-free link poses $\Delta p^l_t$, 4) difference between the current and collision-free link velocities $\Delta v^l_t$, $\Delta \omega^l_t$, and 5) a mask that represents valid feature interpolation regions in the cylindrical projections.

First, we compute the vertex-wise versions of each of these features.
For 1) and 3), this is done by taking the L2 norm of the appropriate vertex locations on the mesh surfaces, transformed by the link poses.
For 2) and 4), this is done by converting linear and angular link velocities into linear vertex velocities: $v^l_t + d^i \otimes \omega^l_t$, where $d^i$ is the difference between the $i$th vertex location and the link's center of mass, $\otimes$ denote the cross product, and angular velocities $\omega$ are expressed with the center of mass as the rotation origin.
For the four of these features, the vertex-wise variants are vectors in $\mathbb{R}^3$.
Second, we take the dot product between the vertex-wise features and each vertex's normal.
Doing this reduces the dimension of vertex-wise features from $3$ to $1$, and it also ensures that they are represented relative to the robot arm's current frame, and not the world frame.
Third, we form a feature image by placing scalar feature values of each vertex onto the corresponding $UV$ coordinates and performing linear interpolation via Delaunay triangles to fill in the rest of the image.
Like the distance field images, these feature images are also $32\times32$. 
Lastly, because not all regions in the $UV$ space correspond to valid vertices on the mesh, we form a mask of valid interpolation regions and zero-out the pixels that are not in the mask.
This mask is also given to the network as an additional channel in the input observations.
We use alpha shapes~\footnote{\url{https://github.com/bellockk/alphashape}} to produce \textit{concave} hulls around the projected mesh vertices and use them as the masks.
See the blue shaded region in panel F of Figure~\ref{fig:mesh} for an illustration of the valid interpolation area.

\subsection{Neural Network Model}

See Figure~\ref{fig:network} for a visualization of the network architecture.
The contact localization model takes in a window of $5$ past observations and predicts the contact state of all robot links at the latest time step in the window.
There are $3$ main modules in the network: a feature extractor, link-specific feature processors, and contact prediction heads.
The feature extractor has $2$ components - a convolutional encoder that processes all the feature images (Figure~\ref{fig:proj}) and a fully-connected submodule to process the non-image observations.
Non-image observations include joint velocities and the difference between the robot's joint angles and those of the robot in the ``collision-free" simulation.
The feature extractor independently processes observations from all links and across all timesteps in the window, concatenating the results into a latent vector.
There are $7$ link-specific feature processors that extract embeddings for each link from the shared latent vector.
Lastly, the contact prediction heads take each link-specific embedding and make the appropriate predictions.
There are two prediction heads: a fully-connected submodule that detects whether or not a link is in contact $b^l$ and a transposed convolution decoder that predicts the contact distance fields to localize detected contacts.

Both the feature extractor and the contact prediction heads are shared across all links, and only the intermediate feature processors are link-specific.
This weight sharing introduces an inductive bias that enables efficient network training.

The loss function is a weighted combination of a mean squared error loss for the contact distance field and a binary cross-entropy loss for contact detections.
Because the dataset is heavily imbalanced (only ~$2$\% positive contacts), the binary cross-entropy loss weighs positive to negative samples with a $50:1$ ratio.
$85$\% of the generated trajectories are used in the training set, with the remaining as the validation set.


To convert the model predictions into contact locations, we first check to see if the contact probability of each arm is over a threshold.
If it is, then we extract contact points from the predicted contact distance fields.
The value of each pixel in the distance field corresponds to the distance to the nearest contact point in pixel-space, so localizing contacts means finding pixels with zero or near-zero values.
Here, we include all pixels below a threshold as the predicted contact points, then we map those pixel coordinates back to points on the mesh surface as the predicted contact locations.

\section{EXPERIMENTS}

We performed two experiments to evaluate our proposed approach.
The first compares the performance of the model that uses transposed convolution to predict contact distance fields (\textbf{CDF}) with that of a model which directly classifies (\textbf{CLS}) dense contact locations.
The second demonstrates applying the trained networks to an obstacle mapping task.
CLS has the same inputs and architecture as CDF, except its output heads, which are replaced with one that directly performs multi-label multi-class classification.
This allows CLS to predict multiple contacts at the same time.
Each class is a vertex of the projection meshes.
Similar classifiers are used by multiple prior works, but our model classifies a total of $4077$ contact points, which is more fine-grained than the $661$ points used by~\cite{zwiener2018contact} and the $20$ by~\cite{popov2020transfer}; both works also use joint torque features, which our model does not.

\subsection{Contact Localization}

\begin{table}[!t]
\centering
\vspace{6pt}
\begin{tabular}{l|lll|ll}
       & \multicolumn{3}{l|}{Contact Detection} & \multicolumn{2}{l}{Contact Localization} \\ \cline{2-6} 
       & ACC               & FNR               & FPR               & AMCD-GT                & AMCD-P                 \\ \hline
CLS    & $\mathbf{94.6}$\% & $31.9$\%          & $4.4$\%  & $\mathbf{0.4}$ ($1.7$) & $5.5$ ($4.6$)          \\
CDF    & $91.9$\%          & $\mathbf{10.5}$\% & $7.8\%$           & $3.6$ ($3.8$)          & $\mathbf{2.3}$ ($3.4$) \\ \hline
CLS-RF & $94.1$\%          & $62.8$\%          & $\mathbf{4.0}$\%           & $0.7$ ($2.0$)          & $3.8$ ($3.9$)          \\
CDF-RF & $78.6$\%          & $11.42$\%         & $21.3$\%          & $4.9$ ($4.4$)          & $2.8$ ($3.6$)
\end{tabular}
\caption{\footnotesize{
    Contact detection and localization results with reduced-feature ablations.
    AMCD units are in cm.
    Parentheses refer to standard deviation.
}}
\label{tab:results}
\end{table}

Two sets of metrics are used to compare contact prediction performance - one for detecting and one for localizing contacts.
For contact detection, we report the accuracy (\textbf{ACC}), the false negative rate (\textbf{FNR}), and the false positive rate (\textbf{FPR}).
Positive means there is a contact.
For evaluating contact localizations, we compute the Average Minimum Contact Distance (AMCD).
Denote a set of ground-truth contact locations on a link expressed in Cartesian space as $[c^l_1, \hdots, c^l_N]$ and the set of predicted contact locations on the same link as $[\hat{c}^l_1, \hdots, \hat{c}^l_M]$.
There are two variants of AMCD, computed as follows:
$\text{AMCD-GT} = \frac{1}{N}\sum_{n=1}^N \min_{m} \|c^l_n - \hat{c}^l_m\|_2$.
$\text{AMCD-P} = \frac{1}{M}\sum_{n=1}^M \min_{n} \|\hat{c}^l_m - c^l_n\|_2$.
A low AMCD-GT means that most ground truth contacts have predicted contacts that are nearby; a low AMCD-P means that most predicted contacts are near ground truth contacts.
AMCD metrics can only be computed for time steps when there are true positives.
Both AMCD and the contact detection metrics need to be viewed together to evaluate model performance.
We also perform an ablation study by removing the image-based features from both CLS and CDF, resulting in models with reduced features, CLS-RF and CDF-RF.

See Table~\ref{tab:results} for results computed for a validation dataset in simulation.
Although CLS has higher ACC and and lower FPR, its FNR is $3\times$ that of CDF.
For localization, CLS has a low AMCD-GT of $0.4$cm, meaning ground truth contacts points mostly have predicted contact points nearby.
However, it has a much higher AMCD-P of $5.5$cm, meaning many of its predicted contact points are far away from ground truth contact points.
By contrast, CDF has similar performance for both AMCD-GT and AMCD-P, achieving $3.6$cm and $2.3$cm respectively.
Because CDF uses transposed convolution layers to predict a distance field, its predicted contacts tend to form clusters, so its errors are more spatially correlated.
Taking the average of the two AMCDs, both CLS and CDF achieve a mean AMCD of $3.0$cm.
While the AMCD degradation of CLS-RF and CDF-RF are apparent but not significant, their contact detection metrics significantly deteriorate --- CLS-RF has double the FNR as CLS, and CDF-RF has almost triple the FPR as CDF.

\subsection{Obstacle Mapping}

\begin{table}[!t]
\centering
\vspace{6pt}
\begin{tabular}{l|ll|ll}
        & \multicolumn{2}{l|}{Simulation} & \multicolumn{2}{l}{Real World} \\ \cline{2-5} 
        & CLS               & CDF             & CLS             & CDF             \\ \hline
ACC     & $95.4$\%          & $95.8$\%        & $96.3$\%        & $95.9$\%        \\
FNR     & $70.3$\%          & $70.7$\%        & $70.5$\%        & $56.3$\%        \\
FPR     & $3.1$\%           & $2.5$\%         & $0.6$\%         & $1.2$\%         \\
NLL     & $0.26$ ($0.54$)   & $0.20$ ($0.24$) & $0.29$ ($0.06$) & $0.27$ ($0.06$) \\
AMVD-GT & $11.6$ ($9.4$)    & $10.5$ ($8.7$)  & $6.5$ ($3.5$)   & $4.6$ ($1.9$)   \\
AMVD-P  & $6.3$ ($5.9$)     & $6.1$ ($5.2$)   & $1.8$ ($1.7$)   & $1.7$ ($1.3$)
\end{tabular}
\caption{\footnotesize{
    Voxel Mapping Results.
    AMVD units are in cm.
    Parentheses refer to standard deviation.
}}
\label{tab:voxels}
\end{table}

\begin{figure}[!t]
    \centering
    \includegraphics[width=\linewidth]{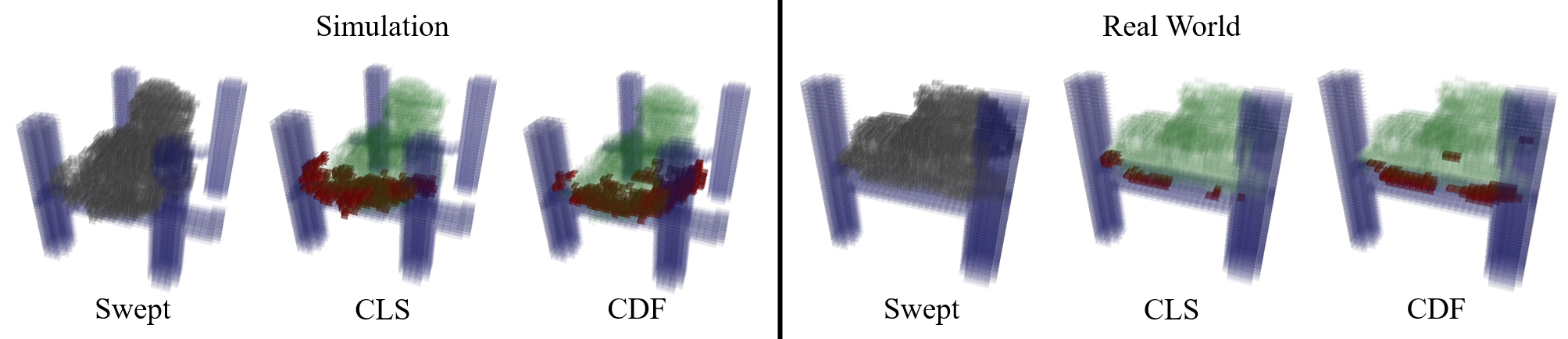}
    \caption{
        Mapped Obstacles in Simulation and Real World.
        The gray voxels in ``swept" are volumes where the robot had explored.
        For all figures, blue are ground truth occupied voxels, red the ones predicted to be occupied, and green predicted to be free.
    }
    \label{fig:voxels}
\end{figure}

We apply the contact prediction models in an obstacle mapping task in both simulation and real world.
Obstacles are modeled with voxel grids with $2$cm resolution.
Voxel values correspond to the probability that a voxel is occupied.
All voxels have initial values of $\frac{1}{K}$, where $K$ is the average number of occupied voxels in the training environments.
To map obstacles, the robot performs a predefined but noisy exploration trajectory.
Contact predictions are used to update the voxels with a Bayesian filter, which treats the probability of occupancy for each voxel to be independent from each other.
We evaluate the negative log-likelihood (NLL), ACC, FNR, and FPR of the voxels in the volume swept by the robot.
We also evaluate a variant of AMCD - the Average Minimum Voxel Distance (\textbf{AMVD}), which computes distances among predicted and ground-truth occupied voxels.
Note these metrics are not about the contact locations on the robot arm, for which we do not have real-world ground truth labels.
Rather, they are for the voxel occupancies of obstacles, which we manually measured in real-world experiments.
Simulation results are aggregated over $200$ noisy exploration trajectories on randomly generated obstacle environments not in the training set.
Real world results use $10$ on one obstacle configuration, also not in the training set.
We control the Franka in the real world with~\cite{zhang2020modular}.

See Table~\ref{tab:voxels} for quantitative results and Figure~\ref{fig:voxels} for a visualization of the mapped voxels.
CLS and CDF achieve comparable performances in both simulation and the real world.
The high voxel FNR is due to the small number of true positives in the volume swept by the robot.
Voxel visualizations show that CDF has less false negative voxels, but slightly more false positive voxels, than CLS.
This is in line with the contact detection results in Table~\ref{tab:results}.

\section{CONCLUSION}
We train a neural network to detect and localize contacts on the surface of a 7-DoF robot arm.
This is done while the robot is moving and without joint torque sensing, relaxing assumptions made in prior works.
A novel cylindrical projection scheme is used to encode features and contact points on mesh surfaces.
The network is trained with domain randomized data in simulation, and we demonstrate its use in an obstacle mapping task in both simulation and real world.
\section{ACKNOWLEDGMENT}
\scriptsize{
The authors thank Saumya Saxena, Brian Okorn, and Tanya Marwah for their insightful discussions.
This work is supported by NSF Grants No. DGE 1745016, IIS-1956163, and CMMI-1925130, the ONR Grant No. N00014-18-1-2775, ARL grant W911NF-18-2-0218 as part of the A2I2 program, and Nvidia NVAIL.
}



\clearpage
\balance
\bibliographystyle{IEEEtran}
\bibliography{citations}

\clearpage
\begin{appendices}
\normalsize

\section{Data Collection}

\subsection{Franka Links}

\begin{figure}[!h]
    \centering
    \includegraphics[width=0.8\linewidth]{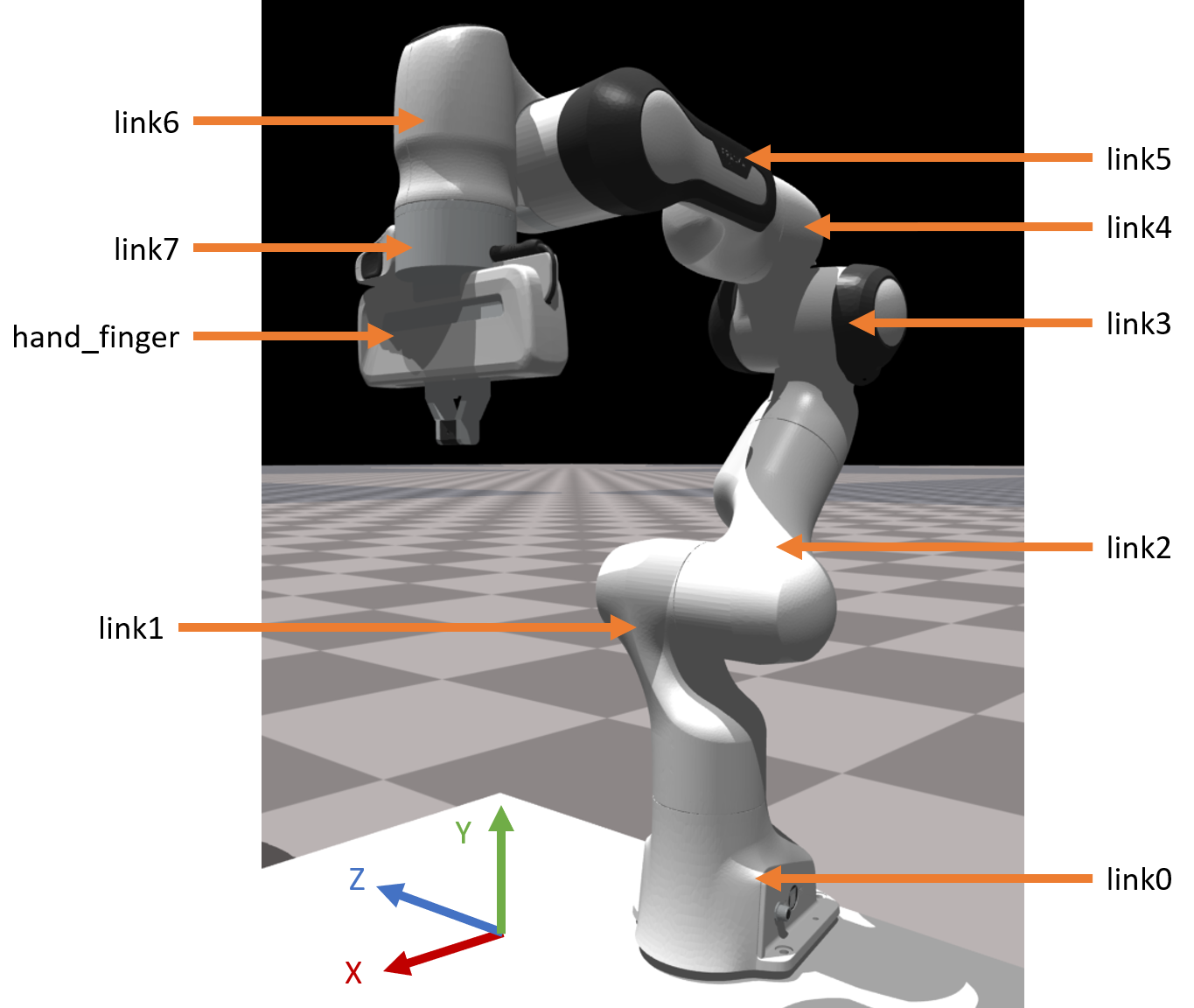}
    \caption{
    Franka Link Names and Coordinate Axes. 
    We collect observations and make predictions for the last $7$ links on the Franka arm, from link2 to hand\_finger, inclusively.
    Franka grippers are closed, and we combine it with the rest of the hand to form the one hand\_finger link mesh.
    }
    \label{fig:link_names}
\end{figure}

\begin{figure*}[!t]
    \centering
    \includegraphics[width=\linewidth]{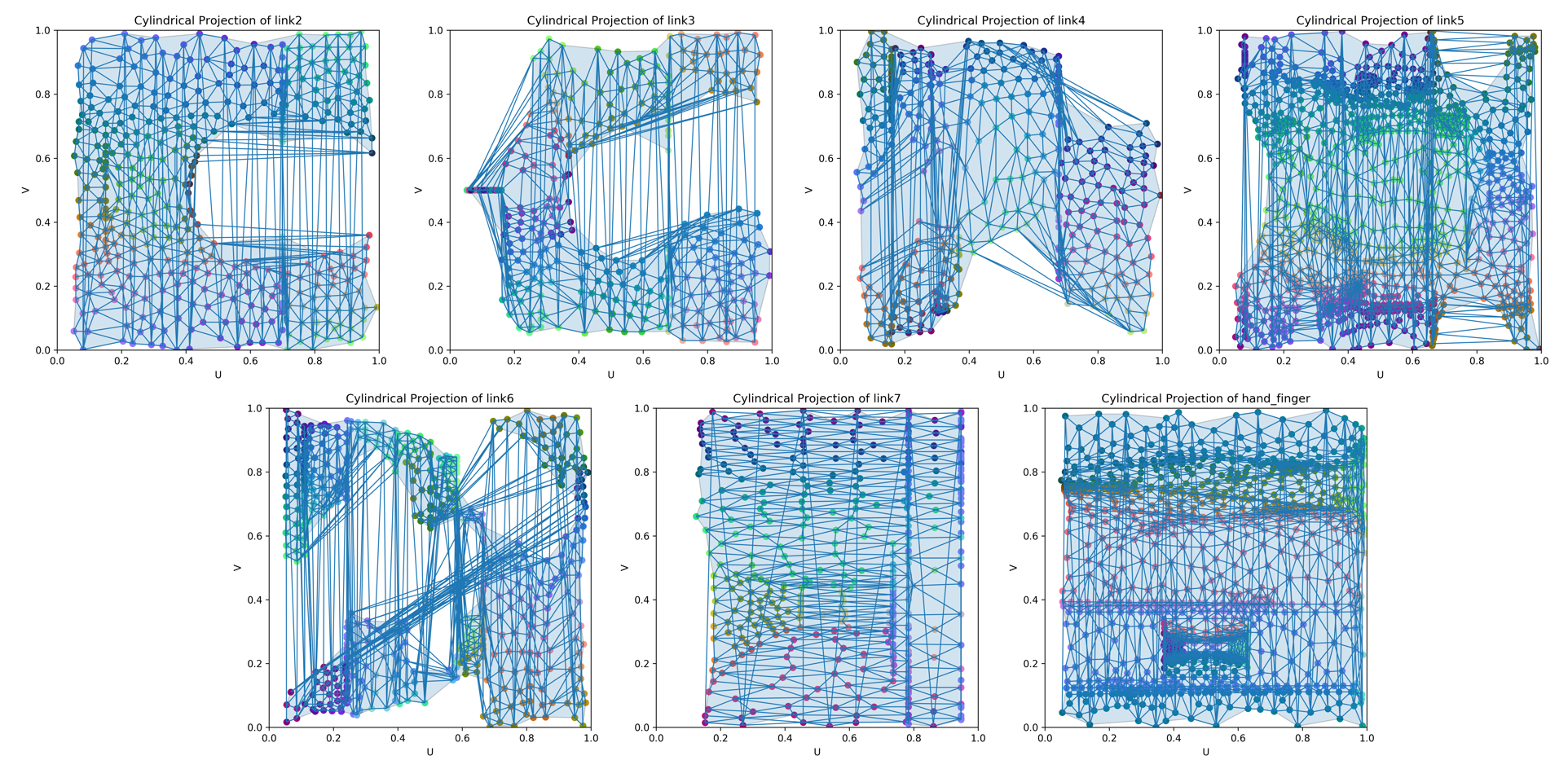}
    \caption{
    Cylindrical Projections for All Links. 
    Cylindrical coordinates are normalized in the range of $[0, 1]$. 
    The horizontal $U$-axis denotes the coordinate along the tube, and vertical $V$-axis denotes the angular coordinate along the tube's circumference. 
    Each dot is a vertex, and the color of the dot indicates its normal direction. 
    Edges are mesh triangle edges. 
    Blue shaded region denotes the concave hull used to define valid projection interpolation region.
    }
    \label{fig:all_cyn_projs}
\end{figure*}

See Figure~\ref{fig:link_names} for the Franka links our trained contact sensing models operate on, and see Figure~\ref{fig:all_cyn_projs} for visualizations of cylindrical projections for all $7$ links.

\subsection{Obstacle Placement}

We generate two sets of random obstacles in the scene.
Each set contains $3$ bars with square cross sections of width $4$cm.
Out of the $3$ bars, $2$ are vertical, and $1$ is horizontal and attached to the vertical bars.
For the vertical bars, we sample an $X$ position uniformly in the range of $[13, 60]$cm in front of the robot.
For the $Z$-axis, one vertical bar samples from $[15, 30]$cm, and the other $[-15, -30]$cm, placing the two bars at both sides of the robot.
Once the positions of the vertical bars are determined, we sample two $Y$ coordinates from the range $[20, 50]$cm.
These values are then used as the points on the vertical bars at which we attach the horizontal bar.

\subsection{Exploration Policies}

For each trajectory, we first sample an initial joint angle configuration within the range $[q_0 - \Delta q, q_0 + \Delta q]$, with $q_0$ being the home configuration seen in Figure~\ref{fig:link_names}, and $\Delta q = [0, 0, 0, 5^\circ, 5^\circ, 10^\circ, 20^\circ]$.
For informed exploration, the robot executes a rectangle-shaped trajectory near where obstacles are typically generated.
Waypoints along the rectangle are perturbed with uniformly sampled noise in $[-3, 3]$cm.
For random exploration, while the trajectory remains under the time horizon of $500$ steps, we sample new delta goal end-effector poses for the robot to reach.
For the translation component of the delta pose, we sample the direction and magnitude separately.
Direction is sampled from a discrete distribution with the following probabilities: 
\begin{align*}
    +X: 0.15, +Y: 0.3, +Z: 0.1\\
    -X: 0.05, -Y: 0.3, -Z: 0.1
\end{align*}
Magnitude is sampled uniformly from the range $[10, 20]$cm.
For the rotation component of the delta pose, we uniformly sample delta euler angles with the range $[0^\circ, 20^\circ]$.
Finally, we uniformly sample a time horizon in the range $[20, 40]$ steps, during which the delta pose command is completed with min-jerk interpolation.

\subsection{Domain Randomization}

In addition to randomizing obstacle placements, we also randomize the inertial parameters of the robot and the gains for the impedance control.
For the inertial parameters, we uniformly sample a mass offset in the range of $[-0.5, 0.5]$kg and and center-of-mass offset in $[-1, 1]$cm, which are added to the base values obtained in~\cite{gaz2019dynamic}.

For the impedance gains, we uniformly sample translation gains $K_T$ in the range $[200, 2000]$ and rotation gains $K_R$ in the range $[3, 6]$.
To use these values, let $K = diag([K_T, K_T, K_T, K_R, K_R, K_R])$ be the $6\times 6$ diagonal gains matrix and $J$ be the $6\times 7$ analytical Jacobian for the robot end-effector that encodes rotations as euler angles.
Then, given a delta pose $p_d$, the commanded torque is $\tau = J^\top (-K p_d - D J \dot{q})$, 
where $D$ is the damping term $D = 2\sqrt{K}$, and $\dot{q}$ is the joint velocity vector.
For simplicity, we have left out the terms that correspond to gravity compensation and Coriolis forces.

\subsection{Data Statistics}

Out of the $2800$ unique trajectories generated, $2400$ were used in the training set.
Each trajectory contains $25$ non-overlapping observation windows of $7$ links, bringing the total amount of training data samples to $2400\times 5\times 7= 420000$.
Out of these, $8358$ samples have at least one positive contact, so there are $1.99\%$ positive contacts in the training dataset.
In addition, $2209$ samples have more than one positive contacts, so there are $0.53\%$ of samples with multiple contacts.

\section{Neural Network Model}

Here we detail the neural network architecture
From Figure~\ref{fig:network}, the Fully-Connected (FC) module of the feature extractor is a multilayer perceptron (MLP) with $3$ hidden layers of sizes $[128, 128, 64]$.
The convolution layers channels $[5, 16, 32, 32]$, each with kernel size $5$ and a stride of $2$.
Each link-specific feature processor is an MLP with hidden layers $[128, 128, 64]$.
The contact distance field decoder has transposed convolution layers with channels $[64, 32, 32, 1]$, kernel sizes $[5, 6, 6]$, and strides of $2$.
The contact detection module is an MLP with hidden layers $[32, 32]$.
We use Leaky-ReLU as the nonlinearities.
Network is implemented with PyTorch Lightning~\cite{falcon2019pytorch}, using the Adam optimizer with a batch size of $256$ and an initial learning rate $1e-3$.

\section{Contact Predictions}

\begin{figure*}[!t]
    \centering
    \includegraphics[width=\linewidth]{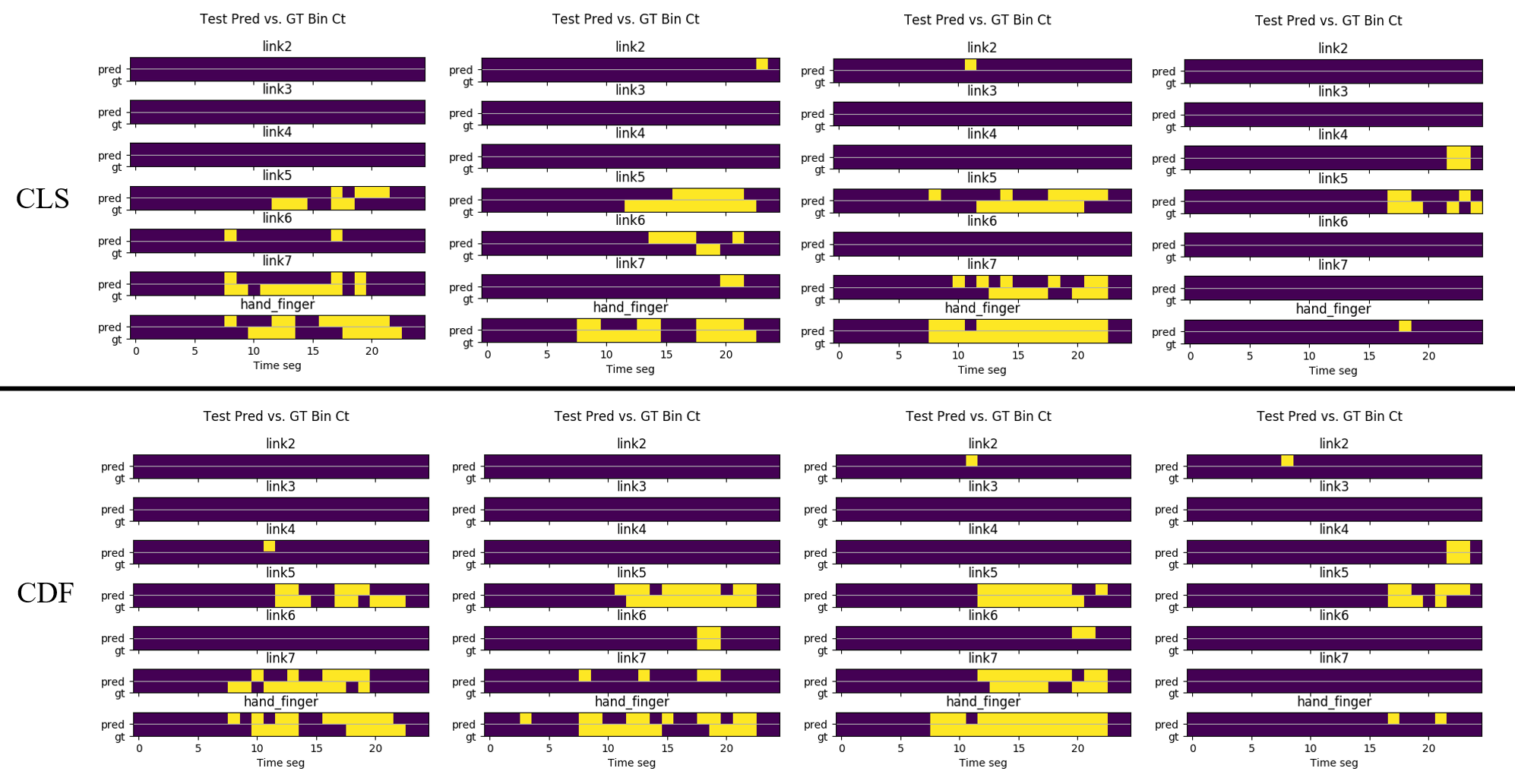}
    \caption{
        Contact Detection Visualizations on Test Data.
        Each column is the execution of one trajectory.
        The top row have plots corresponding to predictions made by CLS, and the bottom CDF.
        In each plot, link-wise ground truth and predicted binary contact detections are visualized, with purple representing negative contacts, and yellow positive contacts.
        The X-axis represents prediction time segments, so each unit corresponds to $5$ observations, one made every $4$ simulation steps.
        In general, CLS has more FNs, and CDF has more FPs, which is consistent with results in Table~\ref{tab:results}.
    }
    \label{fig:trajs}
\end{figure*}

\begin{figure*}[!t]
    \centering
    \includegraphics[width=\linewidth]{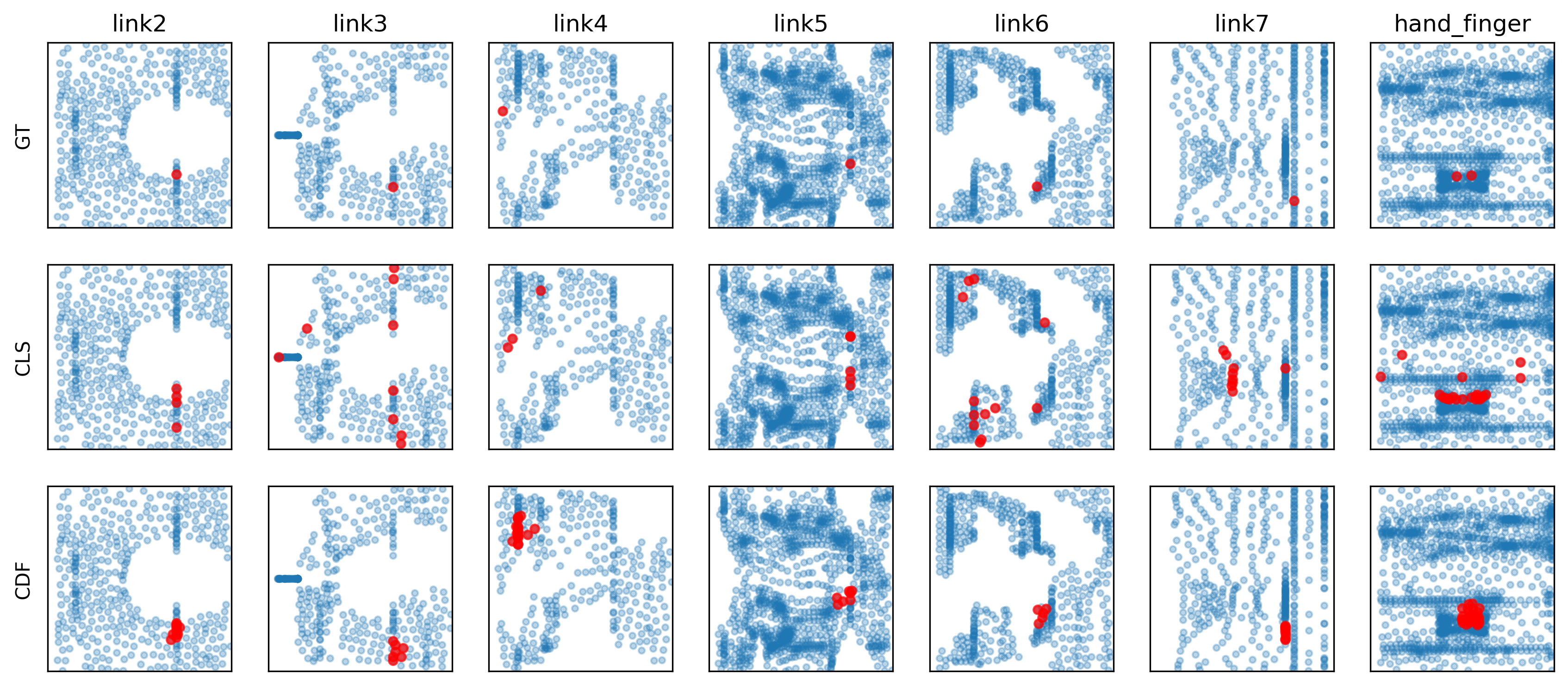}
    \includegraphics[width=\linewidth]{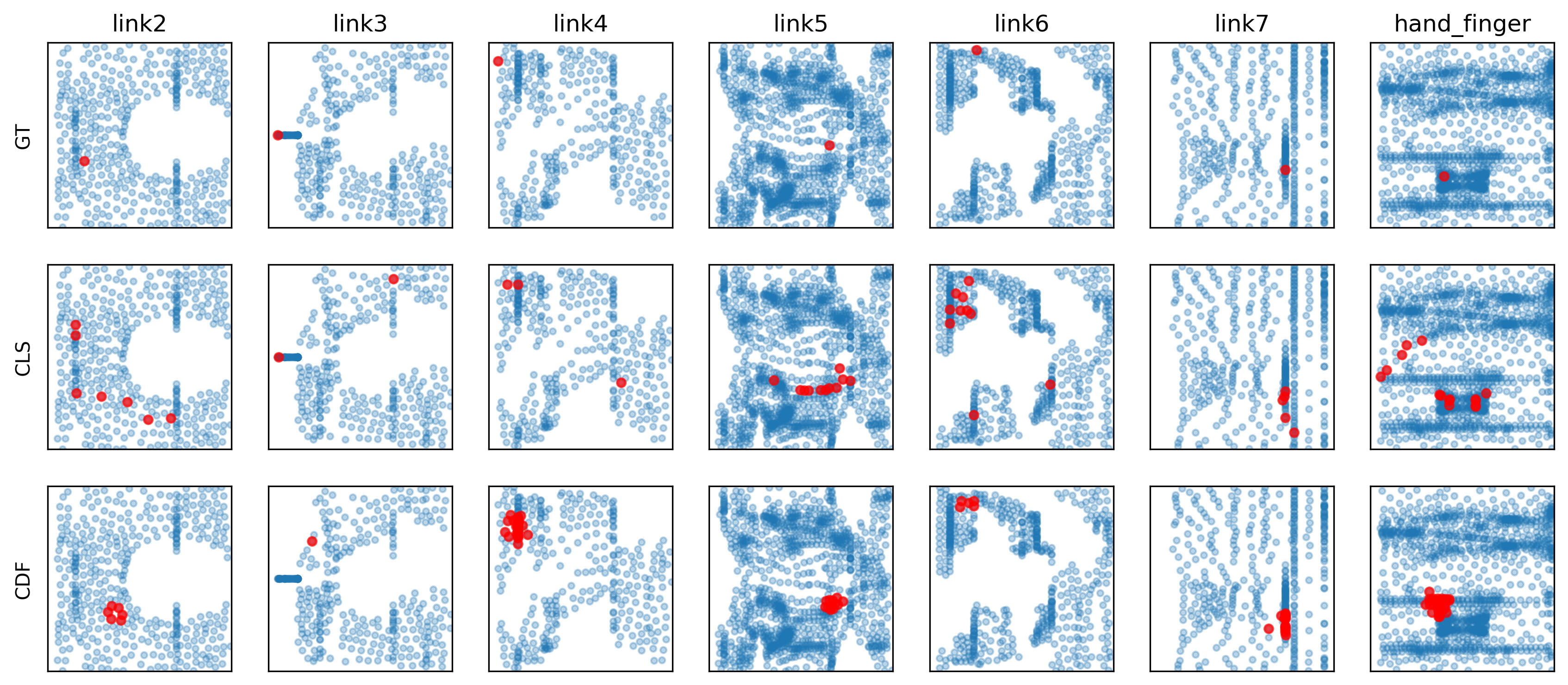}
    \caption{
        Contact Localization Visualizations on Test Data.
        We visualize two sets (top $3$ and bottom $3$ rows) of examples of predicted contact localizations in cylindrical coordinates.
        Blue points are projection mesh vertices; red points are contact locations.
        Each column is a positive sample for a link.
        For each set, the top row are ground truth contact points, middle are predicted by CLS, and bottom by CDF.
        Due to its use of distance fields, CDF's predicted contacts tend to be less scattered than those of CLS.
    }
    \label{fig:locs}
\end{figure*}

See Figure~\ref{fig:trajs} for contact detection visualizations with both CLS and CDF, and Figure~\ref{fig:locs} for contact localization predictions in the cylindrical projection coordinates.

\end{appendices}

\end{document}